# RH-SQL: Refined Schema and Hardness Prompt for Text-to-SQL

Jiawen Yi
Department of Automation
Central South University
Changsha, China
214601043@csu.edu.cn

Guo Chen*
Department of Automation
Central South University
Changsha, China
guo.chen@csu.edu.cn

Zixiang Shen
Department of Automation
Central South University
Changsha, China
zx.shen@csu.edu.cn

*Abstract*—Text-to-SQL is a technology that converts natural language queries into the structured query language SQL. A novel research approach that has recently gained attention focuses on methods based on the complexity of SQL queries, achieving notable performance improvements. However, existing methods entail significant storage and training costs, which hampers their practical application. To address this issue, this paper introduces a method for Text-to-SQL based on Refined Schema and Hardness Prompt. By filtering out low-relevance schema information with a refined schema and identifying query hardness through a Language Model (LM) to form prompts, this method reduces storage and training costs while maintaining performance. It's worth mentioning that this method is applicable to any sequence-to-sequence (seq2seq) LM. Our experiments on the Spider dataset, specifically with large-scale LMs, achieved an exceptional Execution accuracy (EX) of 82.6%, demonstrating the effectiveness and greater suitability of our method for real-world applications.

## I. INTRODUCTION

Text-to-SQL is an artificial intelligence technology that converts natural language questions into Structured Query Language (SQL) code, effectively assisting non-technical users who do not master SQL to bridge the technological divide, enabling rapid functionality for querying structured information. This significantly promotes the reduction of construction costs for information search systems and enhances the efficiency of information tasks. Therefore, Text-to-SQL represents an important research technique in the field of natural language processing currently.

Text-to-SQL technology typically requires input comprising a natural language question reflecting user needs and a schema containing information on existing database tables and columns, as show in Figure 1. Initial studies in this field primarily focused on datasets of lower complexity involving single-turn, single-domain tasks (such as WikiSQL[1]), leading to the proposal of numerous methods (e.g., SQLNet[2], Seq2SQL[1], TypeSQL[3]). However, with advancements in computing power, the rapid development of Large Language Models (LLMs), and increasing societal application demands, more complex datasets have been created, among which single-turn, multi-domain datasets have become the predominant focus of current research. Existing methods

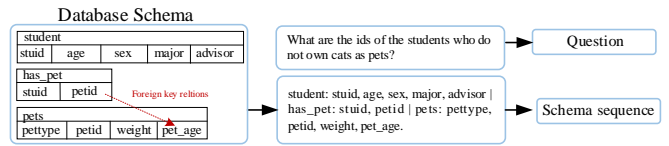

Figure 1.Demonstration of a input case of Text-to-SQL that composed by a natural language question and serialized database schema items (table names and column names)

around these datasets mainly fall into two categories. On one hand, there are methods focused on Schema Linking analysis, which emphasize the modeling and parsing of relationships between schema items, such as LGESQL[4], S2SQL[5], RESDSQL[6], etc. On the other hand, there are methods aimed at evolving Language Models (LMs), focusing on enhancing the structure and understanding capabilities of LMs, such as T5[7], RATSQL[8], RASAT[9], PICARD[10], GAP[11], GRAPPA[12], Graphix-T5[13], etc. These studies have significantly accelerated the development of Text-to-SQL technology.

Additionally, a new research direction has gradually gained attention, based on the decoupling of query difficulty. In Text-to-SQL tasks, the question containing user needs significantly influences the complexity of the corresponding SQL, and in the absence of hints and with varying levels of difficulty, it increases the comprehension pressure on LMs, limiting their performance. The DQHP[14] method identifies the complexity of the corresponding SQL before generating it with the LM and uses multiple LMs to generate SQLs of different complexities, thereby alleviating the comprehension stress on LMs and enhancing the accuracy of SQL generation. However, DQHP requires substantial storage and training costs, making it impractical for real-world applications. Therefore, how to fully utilize LMs while ensuring performance, reducing costs, and facilitating applications has become a key issue in this direction.

To address these concerns, this paper explores Text-to-SQL methods based on query difficulty and proposes the Refined Schema and Hardness Prompt (RH-SQL) method. By using hardness prompts, it alleviates the comprehension pressure on LMs and enhances their performance. Additionally, the model employs a decoupling method to refine schemas, filtering for highly relevant information to improve the accuracy of identifying query hardness and ensuring the quality of hardness prompts.

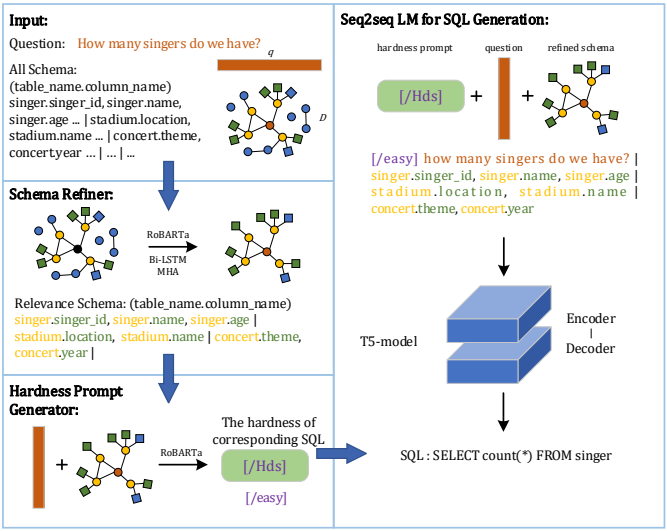

Figure 2. The overview of RH-SQL method. RH-SQL comprises three main components: schema refiner, hardness prompt generator, and language model for SQL generation. Schema contained by table, column and their relations. we demonstrate the use of the hardness prompt in SQL generation.

## II. PROBLEM FORMULATION

### A. Text-to-SQL

The workflow of converting Text-to-SQL involves the following steps: Initially, a natural language question $q$ is presented along with a database $D$ and its corresponding schema $S$. An algorithmic approach is then employed to transform question $q$ into an SQL query $l$. This SQL query $l$, when executed against database $D$, retrieves the answer to question $q$.

### B. Database Schema

The operational context of question $q$ is defined by the relational database $D$. To illustrate, if a database $D$ covers a specific knowledge area $K$, and the query $q$ seeks information outside of $K$, the query is likely to fail. The schema $S$ for database $D$ encompasses three main components: (1) A set of tables $T = \{t_1, t_2, t_3, \ldots, t_n\}$ that organize the data; (2) A set of columns within these tables $C = \{c_1, c_2, \ldots, c_N\}$ that hold the data; and (3) A set of foreign key connections $R = \{(c_{ik}, c_{jh}) | c_{ik}, c_{jh} \in C\}$, which establish the relationships between columns $c_k$ and $c_h$ across different tables $t_i$ and $t_j$, signifying relational links within the data.

### C. SQL Hardness

SQL hardness refers to the difficulty associated with SQL query formulations. According to the Spider benchmark[15], SQL difficulty is segmented into four distinct levels: ***Easy***, ***Medium***, ***Hard*** and ***Extra-hard***. The classification is based on the complexity factors such as the number of SQL elements, the variety of selections, and the intricacy of conditions involved. As a result, SQL statements that incorporate a higher quantity of SQL keywords (for instance, ORDER BY, GROUP BY, INTERSECT, choices of columns, aggregation functions, and the presence of nested subqueries, among others) are considered to be of a higher difficulty level.

## III. RH-SQL METHOD

We propose a method named RH-SQL that alleviates the comprehension stress on LMs and enhances performance through schema refinement and hardness prompts, without making any alterations to the LM. Therefore, RH-SQL is a universally applicable method for Text-to-SQL tasks across sequence-to-sequence (seq2seq) LMs. As show in Figure 2, RH-SQL comprises three main components: Schema Refiner, Hardness Prompt Generator, and Language Model for SQL Generation. The Schema Refiner filters for highly relevant schema information while excluding the less relevant ones to mitigate interference. The Hardness Prompt Generator extracts the hardness level from the input information and provides it to the LM tasked with SQL generation. Finally, an SQL is generated using a fine-tuned seq2seq Pre-trained Language Model (PLM).

### A. Schema Refiner

In Text-to-SQL tasks, the natural language query $q$ and all database schemas $D$ are typically taken as input. However, only a portion of the schema information is relevant to query $q$, and irrelevant information can interfere with the LM's understanding of the task. Herein, we employ a ranking-enhanced encoder as the Schema Refiner, adopted from RESDSQL[6], to identify and filter schema items with high relevance to the question $q$. Specifically, query $q$ and all schema $D$ are input into a model based on an enhanced RoBERTa[16], determining the relevance between schema items (tables and columns) and query $q$, and ranking them accordingly. The top four tables and five most relevant columns, totaling twenty highly relevant schema items, are selected. These 20 items replace the original schema information in the input, forming a new input alongside query $q$ for the Hardness Prompt Generator. Notably, the ranking-enhanced encoder incorporates perception modules designed with Bi-LSTM and multi-head attention into the RoBERTa model, improving the model's perception of tables and columns.

### B. Hardness Prompt Generator

We draw inspiration from DQHP[14]. To identify the complexity of the SQL corresponding to a query $q$ before SQL generation by the LM, thus easing the LM's comprehension stress and enhancing the accuracy of the generated SQL, we fine-tune a RoBERTa-large PLM, known for its proficiency in classification tasks, as the Hardness Prompt Generator. Its primary function is to recognize the complexity level of the SQL code corresponding to query $q$ (*Easy*, *Medium*, *Hard*, *Extra-hard*) by inputting the query $q$ and schema into the PLM. It then generates a hardness prompt that, together with query $q$ and schema, forms the input for the SQL generating LM. The hardness levels Easy, Medium, Hard, and Extra-hard correspond to the prompts '*[/easy]*', '*[/medium]*', '*[/hard]*', and '*[/extra-hard]*', respectively.

## C. Language Model for SQL Generate

Given the versatility of the RH-SQL method, any seq2seq PLM can serve as the SQL generator. In this paper, we employ only a T5 model, a commonly used seq2seq PLM based on the Transformer encoder-decoder architecture, known for its fewer model parameters, strong generalization ability, and adaptability to a wide range of natural language processing problems. The T5 model is the most frequently used language model for Text-to-SQL tasks and comes in various sizes: base, large, and 3b. Since only one LM is required to generate SQL, RH-SQL necessitates merely a quarter of the storage space and training costs compared to DQHP. The hardness prompt, query $q$, and refined schema are input into the T5 model to generate the SQL query tailored to user requirements.

## IV. EXPERIMENT

### A. Dataset and Metrics

In our study, we conduct experiments using the Spider datasets to assess the effectiveness of our approach. The Spider datasets are recognized as among the most demanding benchmarks for evaluating single-turn, multi-domain Text-to-SQL conversion tasks. The structure of the Spider datasets includes a training set, a development set, and an undisclosed test set. The training set is composed of 7,000 instances, distributed across the Easy, Medium, Hard, and Extra-hard difficulty levels at percentages of 24.2%, 39.67%, 20.87%, and 15.26%, respectively. Meanwhile, the development set features 1,034 instances, with the distribution across the difficulty levels being 23.98%, 43.13%, 16.83%, and 16.05%, respectively. The test set, consisting of 2,147 instances, remains unpublished. Therefore, our experiments are solely based on the development set provided by Spider.

For evaluating the effectiveness of the Text-to-SQL parser, we utilize two evaluation metrics as introduced by Yu et al. [15] and Zhong, Yu, and Klein [17]: Exact-set-Match accuracy (EM) and Execution accuracy (EX). The EM metric measures the accuracy with which the predicted SQL query matches the intended (gold) SQL query in terms of its conversion into a certain data structure [15]. On the other hand, the EX metric assesses the congruence between the outcomes of executing the predicted SQL query and those of the gold SQL query, thereby accounting for the accuracy of produced values.

Our experiments were conducted using two local devices equipped with NVIDIA GeForce RTX 3090Ti (24GB) GPUs.

### B. Performance on Spider

In this paper, we aim to demonstrate the efficacy of the RH-SQL method in Text-to-SQL tasks by comparing RH-SQL with five different T5-based Text-to-SQL methods across various model sizes. As shown in Table I, at the base level, RH-SQL-base significantly enhances performance over T5-base[10], with a 13.1% improvement in EM and a 19.9% increase in EX. Moreover, at the large scale, RH-SQL-large achieves a 10.9% increase in EM and a 13.0% increase in EX over T5-large. This indicates that RH-SQL effectively alleviates the comprehension stress of LMs on Text-to-SQL tasks, thereby significantly improving model performance. In practice, the size of an LM determines its understanding capabilities, with larger-scale LMs typically performing better. However, it is noteworthy that RH-SQL-large, when compared with four other 3B-scale methods (which are approximately 12 times the size of the base-scale model, such as T5-3B[10], RAT-SQL[8], RASAT[9], Graphix-T5[13]), still leads in EX performance. This demonstrates that RH-SQL makes full use of language models, breaking the limitations imposed by model size. Furthermore, we enhanced RH-SQL with an intermediary representation method, NatSQL[18], to further boost its performance. Compared to RESDSQL-large of the same scale, our method achieved an 0.7% improvement in EX. Due to limited computing capabilities, we did not conduct experiments with RH-SQL at the 3B scale in this paper. However, its superior performance over 3B methods highlights RH-SQL's practical value with lower computational demands, making it more suitable for engineering applications.

TABLE I. THE EM AND EX RESULTS ON SPIDER'S DEVELOPMENT SET (SPIDER$_D$) WITH LANGUAGE MODEL SIZE FOR SQL GENERATION (#SIZE) (%).

| Method | Spider$_D$ | | |
|---|---|---|---|
| | *#size* | *EM* | *EX* |
| T5-base[10] | ×1 | 57.2 | 57.9 |
| T5-large[10] | ×3 | 65.3 | 67.0 |
| T5-3B[10] | ×12 | 71.5 | 74.4 |
| T5-large+PICARD[10] | ×3 | 69.1 | 72.9 |
| T5-3B+PICARD[10] | ×12 | 75.5 | 79.3 |
| RAT-SQL+GAP+NatSQL[8] | ×12 | 73.7 | 75.0 |
| RASAT[9] | ×12 | 72.6 | 76.6 |
| Graphix-T5-3B[13] | ×12 | 75.6 | 78.2 |
| RESDSQL-large+NatSQL[6] | ×3 | 76.7 | 81.9 |
| DQHP-large [14] | ×12 | 76.5 | 80.4 |
| RH-SQL-base | ×1 | 70.3 | 77.8 |
| RH-SQL-base+NatSQL | ×1 | 73.6 | 80.4 |
| RH-SQL-large | ×3 | 76.2 | 80.0 |
| RH-SQL-large+NatSQL | ×3 | 75.3 | **82.6** |

TABLE II. THE STORAGE AND TRAINING TIMES RECORDS OF RH-SQL AND DQHP ON LARGE-SCALE LEVEL.

| Method | *Storage* | *Training Time* |
|---|---|---|
| DQHP | 15.0GB | 102800s |
| RH-SQL | 6.2GB | 49920$s$ |

TABLE III. THE CLASSIFICATION ACCURACY OF HARDNESS PROMPT GENERATOR (HPG) WITH AND WITHOUT SCHEMA REFINER (SR) (%).

| Method | Classification Accuracy | | | | |
|---|---|---|---|---|---|
| | *Easy* | *Medium* | *Hard* | *Extra* | *All* |
| HPG | 92.89 | 83.98 | 68.26 | 84.44 | 83.56 |
| HPG w/o SR | 94.89 | 69.42 | 57.39 | 67.56 | 70.50 |

## C. Storage Costs Analysis

To validate that our proposed method is more storage and training efficient compared to existing methods based on the SQL queries complexity, we compared DQHP-large with RH-SQL-large in terms of storage space and training time on Spider training set (with 4 batch size and 64 epochs), as detailed in Table II. Due to DQHP's use of a distributed generator and a two-stage training strategy, its storage and training costs are substantial. In contrast, RH-SQL saves 8.8GB in storage space and reduces training time by approximately 50%. This significant reduction in costs greatly enhances the convenience and efficiency of using our approach and its applicability in real-world scenarios.

## D. Hardness Prompt Generator Result and Analysis

To analyze the effect of filtering irrelevant information by the Schema Refiner on the improvement of the Hardness Prompt Generator, we conducted an ablation study. This study compares the classification accuracy of the Hardness Prompt Generator with and without the Schema Refiner, as detailed in Table III. As can be observed from Table III, the overall accuracy increased by 13.06% after incorporating the Schema Refiner. While the accuracy for the Easy category slightly decreased, the Medium, Hard, and Extra-hard categories experienced significant accuracy improvements of 14.56%, 10.87%, and 16.88%, respectively. Therefore, the role of the Schema Refiner in filtering superfluous information is critically important for ensuring the accuracy of the Hardness Prompt Generator.

## E. Limitations

While our proposed method reduces storage and training costs without compromising performance, the classification accuracy for the Hard category within the hardness prompt generator is only 68.26%. This limitation restricts the effectiveness of the hardness prompt in assisting the SQL generator LM. Therefore, there is significant room for improvement in the classification method for hardness.

## V. CONCLUSION

In this paper, we contemplate the issue of high storage and training costs associated with existing algorithms, which are not conducive to practical application, through the lens of research methods based on SQL query complexity. Consequently, we propose an approach for Text-to-SQL that can be embedded into any seq2seq LM, based on Refined Schema and Hardness Prompt. By analyzing the relevance between schema and queries to filter out irrelevant schema information, we enhance the efficiency of the input data. Moreover, by identifying SQL hardness using RoBERTa with the query and highly relevant schema information to form a hardness prompt, we provide cues for the SQL generator, thereby reducing the LM's storage requirements and training costs. Experiments conducted on the Spider dataset demonstrate that RH-SQL significantly improves the LM's understanding of Text-to-SQL tasks and achieves commendable performance, making it more suitable for real-world applications compared to existing methods based on SQL query complexity.

ACKNOWLEDGMENT

This work is supported by National Naturel Science Foundation of China under Grant No. 62073344.